\documentclass[conference]{IEEEtran}
\IEEEoverridecommandlockouts

\usepackage{cite}
\usepackage{amsmath,amssymb,amsfonts}
\usepackage{algorithmic}
\usepackage{graphicx}
\usepackage{textcomp}
\usepackage{xcolor}

\usepackage{booktabs} 
\usepackage{multirow}
\usepackage{float}

\def\BibTeX{{\rm B\kern-.05em{\sc i\kern-.025em b}\kern-.08em
    T\kern-.1667em\lower.7ex\hbox{E}\kern-.125emX}}
\begin{document}

\title{Adaptive Interactive Segmentation for Multimodal Medical Imaging via Selection Engine

}

\author{
Zhi Li\textsuperscript{1}, 
Kai Zhao\textsuperscript{2}, 
Yaqi Wang\textsuperscript{3}\textsuperscript{\textasteriskcentered}, 
Shuai Wang\textsuperscript{1}\textsuperscript{\textasteriskcentered} \\
\textsuperscript{1}\textit{Hangzhou Dianzi University}, Hangzhou, China \\
\textsuperscript{2}\textit{Department of Neurosurgery, First Medical Center, Chinese PLA General Hospital}, Beijing, China \\
\textsuperscript{3}\textit{Communication University of Zhejiang}, Hangzhou, China \\
\vspace{-2.5em} 
\thanks{\hrule\vspace{0.8em}\textsuperscript{\textasteriskcentered}Corresponding authors: Yaqi Wang @wangyaqi@cuz.edu.cn and Shuai Wang @shuaiwang.tai@gmail.com.}
}


\maketitle


\begin{abstract}

In medical image analysis, achieving fast, efficient, and accurate segmentation is essential for automated diagnosis and treatment. Although recent advancements in deep learning have significantly improved segmentation accuracy, current models often face challenges in adaptability and generalization, particularly when processing multi-modal medical imaging data. These limitations stem from the substantial variations between imaging modalities and the inherent complexity of medical data. To address these challenges, we propose the Strategy-driven Interactive Segmentation Model (SISeg), built on SAM2, which enhances segmentation performance across various medical imaging modalities by integrating a selection engine. To mitigate memory bottlenecks and optimize prompt frame selection during the inference of 2D image sequences, we developed an automated system, the Adaptive Frame Selection Engine (AFSE). This system dynamically selects the optimal prompt frames without requiring extensive prior medical knowledge and enhances the interpretability of the model’s inference process through an interactive feedback mechanism. We conducted extensive experiments on 10 datasets covering 7 representative medical imaging modalities, demonstrating the SISeg model’s robust adaptability and generalization in multi-modal tasks. The project page and code will be available at:https://github.com/RicoLeehdu/SISeg.
\end{abstract}

\section{Introduction}
Medical image segmentation plays a critical role in clinical practice, supporting essential tasks such as diagnosis, treatment planning, and disease monitoring\cite{b1}. Accurate segmentation of tissues, organs, and lesions from medical images not only improves diagnostic precision but also enhances treatment planning\cite{b2}. However, manual segmentation is a labor-intensive, time-consuming process requiring substantial expertise. In clinical settings, the extensive manual annotation effort renders segmentation inefficient and costly\cite{b3}. To overcome these limitations, semi-automatic and fully automatic segmentation methods have emerged, aiming to improve segmentation consistency and efficiency by minimizing manual intervention, thus facilitating the analysis of large-scale medical imaging data\cite{b4}.  

In recent years, deep learning models have propelled interactive segmentation methods to the forefront, particularly in scenarios that require real-time feedback and efficient annotation\cite{b4}. By allowing users to provide minimal key cues, such as bounding boxes around lesions or positive and negative sample points, models can automatically propagate segmentation results without extensive labeling\cite{b5,b6,b8}. This is particularly beneficial in medical image segmentation, where real-time interaction can significantly boost clinician productivity, reduce labeling costs, and enhance performance in sequence data processing\cite{b9}.  

While the Segment Anything Model 2 (SAM2)\cite{b11} has demonstrated impressive generalization and interactive segmentation capabilities in natural image tasks, it faces significant challenges when applied to medical imaging\cite{b29,b30}. Medical images often exhibit complex textures, variable contrasts, and artifacts, making it difficult for a single prompt type to generalize effectively across modalities\cite{b9}. Moreover, medical image segmentation frequently involves processing large-scale 2D or 3D sequences\cite{b7}, where conventional models struggle to balance segmentation accuracy, inference efficiency, and memory consumption\cite{b10}.

To address these challenges, we propose a strategy-driven intelligent interactive segmentation system (SISeg), designed to adaptively handle multi-modal medical image segmentation. SISeg integrates multiple prompt types and builds upon the Segment Anything Model 2 framework. At the core of SISeg is the Adaptive Frame Selection Engine (AFSE), which dynamically selects the most appropriate prompt frames based on image characteristics, without relying on prior medical knowledge. This engine not only reduces memory consumption but also enhances interpretability in the segmentation process, particularly for sequential data. By incorporating an unsupervised scoring mechanism, SISeg effectively processes diverse modalities such as dermoscopy, endoscopy, and ultrasound, achieving superior segmentation accuracy even in complex scenarios.
This work presents three main contributions:
\begin{itemize}
    \item A novel, flexible multi-prompt segmentation framework (SISeg) that efficiently handles diverse medical imaging modalities without requiring extensive domain-specific knowledge.
    \item The introduction of the Adaptive Frame Selection Engine (AFSE), which dynamically optimizes prompt frame selection, significantly improving inference efficiency while reducing memory usage.
    \item Extensive experiments on 10 datasets across 7 distinct medical imaging modalities, validating the superior generalization capability and reduced annotation burden of the SISeg framework.
\end{itemize}

\section{RELATED WORK}
\subsection{Segment Anything Model in Medical Imaging}
Medical foundation models are large-scale, pre-trained models designed for rapid customization through fine-tuning or in-context learning\cite{b10,b12,b13,b14}. Despite substantial progress, challenges remain in tasks such as image segmentation, largely due to the scarcity of annotated masks\cite{b9}. These limitations hinder SAM's performance, particularly in cross-modal segmentation tasks, where the impact of different prompt types has not been thoroughly investigated\cite{b12}. Approaches like MedSAM\cite{b9} and Medical SAM Adapter\cite{b15} have attempted to enhance cross-modal segmentation by fine-tuning SAM\cite{b5} using bounding box prompts on large medical datasets. However, these methods remain constrained by their focus on a limited set of prompt types across modalities. In response to these challenges, we propose a segmentation strategy that integrates multiple prompt types and evaluate its effectiveness across seven representative medical imaging modalities.

\subsection{Medical Image Segmentation}
Medical image segmentation is fundamental in modalities such as CT, MRI, and ultrasound\cite{b17,b18,b19}. Models like U-Net\cite{b20} and its variants\cite{b21,b22,b23} have demonstrated exceptional performance in this domain. However, most models lack cross-modal robustness, struggle with large-scale datasets, and require extensive manual tuning, failing to optimize prompt selection or interaction efficiency\cite{b26,b27}. To address these challenges, we introduce an automated prompt selection framework that intelligently selects prompt frames without relying on domain-specific medical knowledge, significantly reducing memory usage and enhancing inference efficiency, especially for sequence data.

\section{METHODOLOGY}
\subsection{Strategic Interactive Segmentation System}
\textbf{Preliminaries.} 
As shown in Fig. \ref{fig:fig1}, SAM2\cite{b11} integrates an image encoder, memory encoder, and memory attention mechanism to enhance segmentation by leveraging both current and historical frame information. The image encoder abstracts the input into an embedded representation, while the memory encoder processes representations from previous frames. The memory attention mechanism integrates historical information to refine segmentation of the current frame. This architecture employs a hierarchical visual Transformer as the encoder and a lightweight bidirectional Transformer as the decoder, combining cue and image embeddings. 

In this work, we introduce two key modules to optimize the interactive segmentation process: an unsupervised scoring mechanism (Scorer) and a Selector, which aids in selecting representative frames, as shown in Fig. \ref{fig:fig1}.

\subsection{Exploring Robust Prompts}
Prompt-based segmentation, particularly with pre-trained models, has become a standard approach for reducing labeling costs in related tasks. SAM2 extends prompt-based segmentation to video contexts and has demonstrated strong performance in 3D medical imaging, such as CT and MRI, by treating volumetric data as video streams. However, its potential in 2D medical image segmentation remains underexplored.

We found that SAM2 can effectively apply both One-Prompt and Multi-Prompt segmentation across sets of 2D medical images, treating them as video sequences. In this approach, the model requires only a few well-chosen image prompts to achieve segmentation across the entire image set—a task that proves challenging for other methods. Therefore, we investigated the effectiveness of various prompt types across different medical imaging modalities. The specific prompt types and examples are illustrated in Fig. \ref{fig:fig1}.

\subsection{Adaptive Frame Selection Engine}
Single-point segmentation\cite{b28} allows users to provide a single hint to the model for an unseen example. This approach capitalizes on the model’s generalization ability without requiring retraining or fine-tuning. While this technique is effective for tasks such as optic disc and cup segmentation in fundus images, it may not generalize well across all medical imaging modalities due to the reliance on easily recognizable prior knowledge in some tasks.

\begin{figure*}[htbp]
    \centering 
    \includegraphics[width=\textwidth]{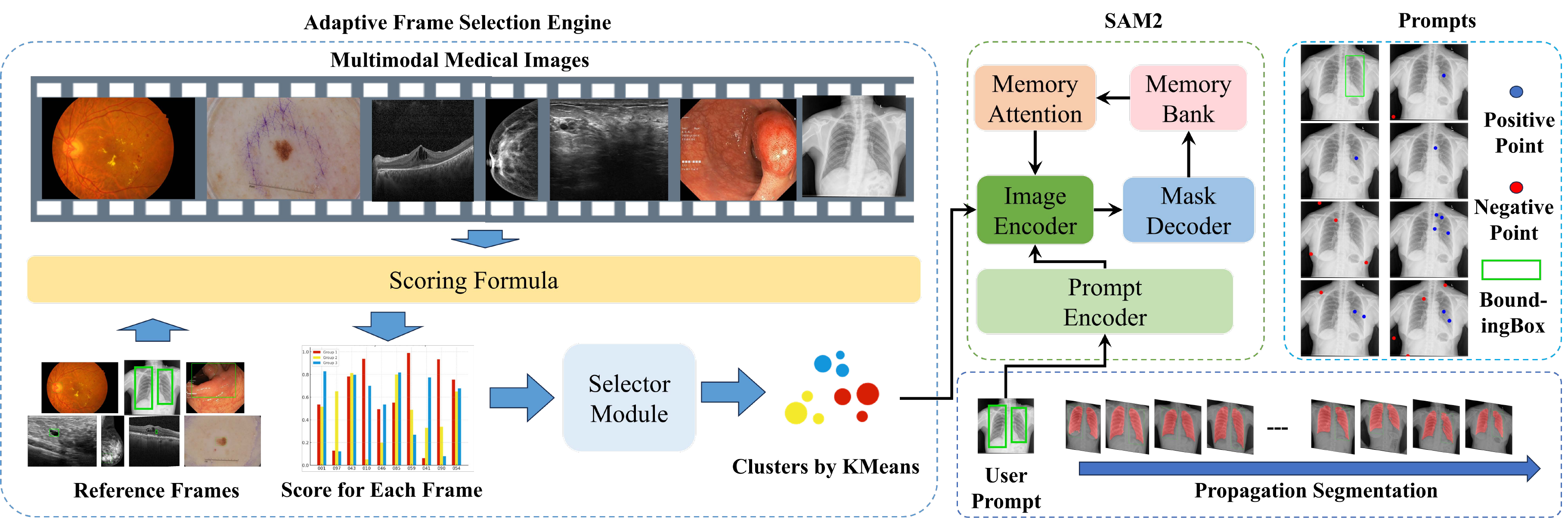} 
    \caption{The model structure of SISeg. The figure illustrates the SAM2 architecture, which includes the image encoder, prompt encoder, memory attention mechanism, streaming memory bank, and mask decoder. AFSE leverages reference frames provided by clinicians to enable fully automated feedback and segmentation propagation. The top-right section highlights an example with chest X-rays, demonstrating the application of various prompt types, such as true negative point prompts and bounding box prompts, for model-based segmentation.}
    \label{fig:fig1}
\end{figure*}

\textbf{Scoring Formula.}
To address this limitation, we propose an unsupervised scoring mechanism that evaluates the dataset based on image features, aiding in the selection of representative frames for annotation. Let $B$ represent the brightness score, $C$ the contrast score, $E$ the edge density, $H$ the color histogram similarity, and $S$ the shape similarity. These variables are combined to form a composite score $F$, which is calculated for each image relative to a reference frame in the dataset. The overall composite score is defined as:

\begin{equation}
F = \alpha \cdot B + \beta \cdot C + \gamma \cdot E + \delta \cdot H + \epsilon \cdot S
\end{equation}

Where:
\begin{itemize}
    \item $\alpha$, $\beta$, $\gamma$, $\delta$, and $\epsilon$ are weights assigned to each feature score.
    \item $B$ (brightness) is computed as the mean brightness of the grayscale image, normalized to the range [0, 1]:
    \begin{equation}
    B = \frac{\text{mean}(I_{\text{gray}})}{255}
    \end{equation}
    \item $C$ (contrast) is defined as the standard deviation of the grayscale pixel intensities:
    \begin{equation}
    C = \frac{\text{std}(I_{\text{gray}})}{255}
    \end{equation}
    \item $E$ (edge density) represents the proportion of edge pixels, computed using the Canny edge detector:
    \begin{equation}
    E = \frac{\text{mean}(I_{\text{edges}})}{255}
    \end{equation}
    \item $H$ (color histogram similarity) measures the correlation between the HSV histograms of the current image and the reference frame:
    \begin{equation}
    H = \text{corr}\left(\text{hist}_{\text{HSV}}(I), \text{hist}_{\text{HSV}}(I_{\text{ref}})\right)
    \end{equation}
    \item $S$ (shape similarity) is derived from the Hu moments of the grayscale image and the reference frame, defined as:
    \begin{equation}
    S = -\log\left(\sum_{i=1}^{7} \left| M_{\text{Hu}}(I_{\text{ref}})_i - M_{\text{Hu}}(I)_i \right| + \epsilon \right)
    \end{equation}
    where $M_{\text{Hu}}$ represents the Hu moments, and $\epsilon$ is a small constant to prevent division by zero.
\end{itemize}

The composite score $F$ is calculated for each image, and K-means clustering is applied to group frames into clusters based on their similarity to the reference frame.

\textbf{Selector Module.}
The interaction engine incorporates a selector module to enable efficient dataset navigation. The Selector initially identifies a reference frame $I_{\text{ref}}$, chosen by the clinician for its clinical relevance. Each frame in the dataset is then assigned a composite score $F$, computed based on its similarity to the reference frame. This composite score integrates various features, such as brightness, contrast, edge density, color histogram similarity, and shape similarity, as defined in Equation (1). The process is illustrated in Fig. \ref{fig:fig1}.

To group similar frames, we apply the KMeans clustering algorithm. Given a set of $N$ frames, let $\mathbf{X} = \{F_1, F_2, ..., F_N\}$ represent the composite scores of these frames. The KMeans algorithm minimizes the following objective function:

\begin{equation}
\min_{C_1, C_2, ..., C_k} \sum_{i=1}^{k} \sum_{x \in C_i} \| x - \mu_i \|^2
\end{equation}

where $C_i$ is the $i$-th cluster, and $\mu_i$ is the centroid of $C_i$. The algorithm iteratively minimizes the sum of squared distances between each frame’s composite score and its cluster centroid. After clustering, frames closest to the centroids are selected as representative frames for segmentation. The Scorer module then ranks the remaining frames by their proximity to these centroids, providing feedback to the user on the most relevant frames for further annotation.

Subsequently, SISeg automatically propagates segmentation based on the prompts associated with the selected key frames. Figure \ref{fig:fig1} demonstrates an example of lung segmentation in X-ray mode. This process reduces manual intervention and enhances segmentation efficiency by partitioning long sequence datasets into clusters based on scoring, thereby lowering inference costs.

\section{EXPERIMENTS}
\subsection{Experimental Setup}

\textbf{Dataset.} In our experiments, we utilized seven distinct medical imaging modalities: Dermoscopy (Der), Endoscopy (Endo), Fundus, Optical Coherence Tomography (OCT), Ultrasound (US), X-ray (XRay), and Mammography (MG). Ten publicly available datasets were employed, including: PAPILA \cite{PAPILA}, Breast Ultrasound \cite{BUS-BRA}, Kvasir-SEG \cite{Kvasir-SEG}, IDRiD \cite{IDRiD}, ISIC 2018 \cite{ISIC2018}, Intraretinal Cystoid Fluid \cite{IntraretinalCystoidFluid}, CDD-CESM \cite{CDD-CESM}, m2caiSeg \cite{maqbool2020m2caiseg}, Chest X-ray Masks and Labels \cite{ChestXrayMasks}, and hc18 \cite{van2018automated}. Due to inference cost constraints, the datasets were split into training and validation sets using a 7:3 ratio with scikit-learn \cite{scikit-learn}, applying a fixed seed of 2024 to ensure consistency. The validation set comprises 30\% of the original data.

\textbf{Implementation Details.}  
The experiments were conducted on a single RTX 3090 GPU. We employed four pre-trained SAM2 model variants: sam2 hiera tiny, sam2 hiera small, sam2 hiera base plus, and sam2 hiera large, to effectively adapt the model to the medical imaging datasets. The evaluation metrics used for performance assessment were Dice coefficient and Intersection over Union (IoU).

\subsection{Results and Analysis}

\begin{figure}[H]
    \centering 
    \includegraphics[width=\linewidth]{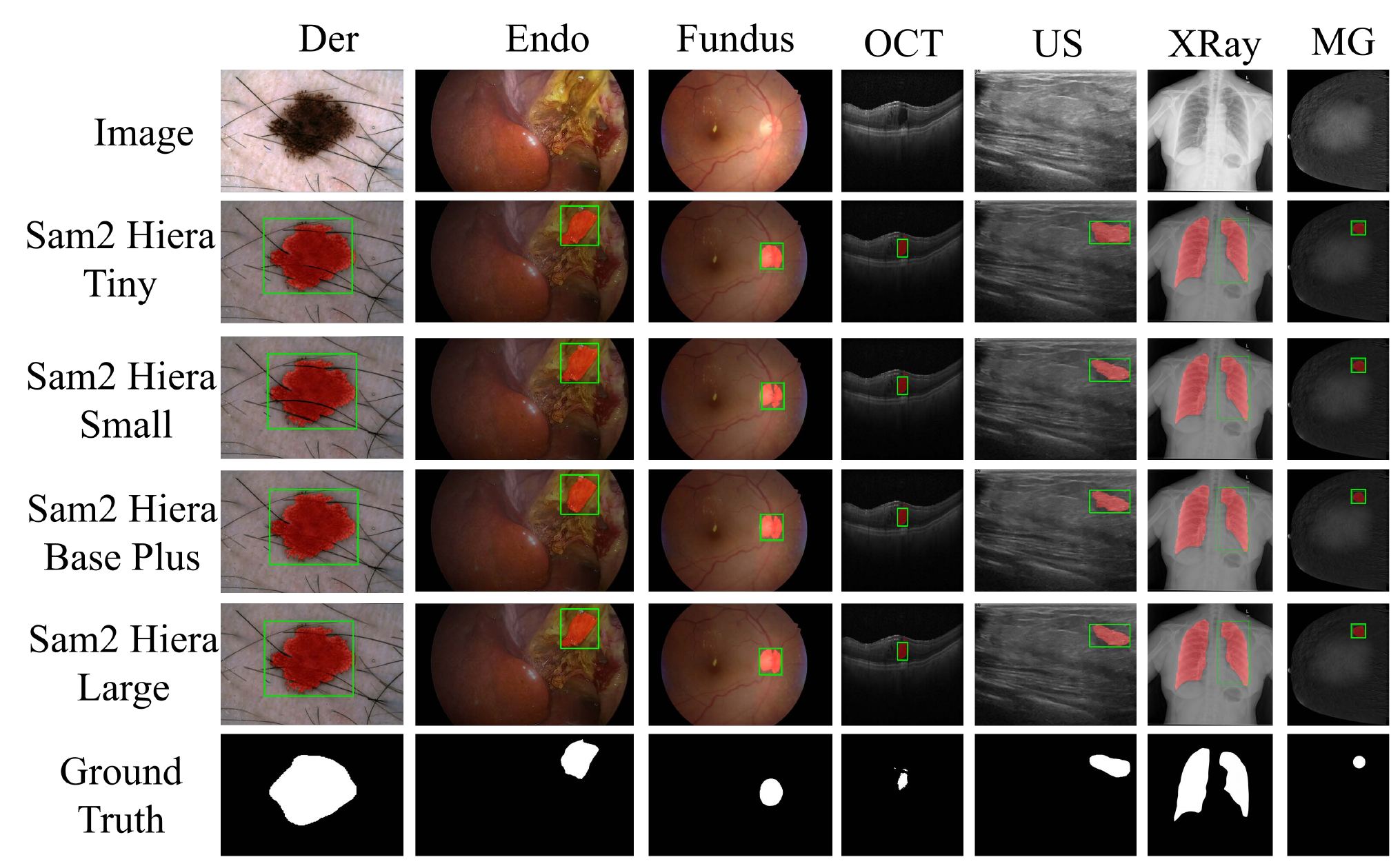} 
    \caption{Visualization of zero-shot segmentation results produced by different SAM2 Hiera models across various medical imaging modalities.}
    \label{fig:fig2}
\end{figure}

\textbf{SAM2 Zero-Shot Performance.}  
Table \ref{tab:tab1} presents the segmentation performance of various SAM2 Hiera models, evaluated using the Bounding Box prompt across different medical imaging modalities. The SAM2 variants exhibited strong zero-shot segmentation performance, with each variant performing optimally in specific modalities, as illustrated in Fig. \ref{fig:fig2}.

\textbf{Effects of AFSE.}  
The Adaptive Frame Selection Engine (AFSE) consistently outperformed both random and uniform strategies across multiple modalities, as shown in Table \ref{tab:tab2}. AFSE's scoring mechanism ensures that the selected frames are more representative and clinically relevant, leading to improved segmentation accuracy and reduced annotation effort.

Notably, AFSE surpasses the second-best method, AFSE (without scorer), by 9.39\% in X-ray and 10.97\% in Mammography, underscoring the significance of the scoring mechanism. Additionally, AFSE demonstrates a 6.33\% improvement in Endoscopy and a 10.66\% improvement in OCT.

\begin{table}[t]
\centering
\caption{Comparison of Dice and IoU scores across different SAM2 Hiera model types for various medical imaging modalities.}
\label{tab:tab1}
\resizebox{\columnwidth}{!}{
\begin{tabular}{c|c|ccccccc}
\hline
Model Type & Metric & Der & Endo & Fundus & OCT & US & XRay & MG \\
\cline{3-9}
\hline
SAM2 Hiera  & Dice & \textbf{94.79} & 85.27 & 99.92 & 90.31 & 94.44 & \textbf{96.55} & 81.56 \\  \cline{2-9}
 Tiny       & IoU  & \textbf{90.98} & 77.64 & 99.88 & 83.96 & 90.19 & \textbf{94.01} & 75.18 \\ \hline
SAM2 Hiera  & Dice & 91.84 & 85.48 & 99.96 & 90.25 & 93.73 & 96.47 & 81.26 \\ \cline{2-9}
 Small      & IoU  & 87.39 & 78.03 & 99.94 & 83.93 & 89.59 & 93.93 & 74.63 \\ \hline
SAM2 Hiera  & Dice & 94.11 & 86.24 & 99.98 & \textbf{90.39} & 94.51 & 95.51 & 80.72 \\ \cline{2-9}
 Base Plus  & IoU  & 90.04 & 78.59 & 99.96 & \textbf{84.13} & 90.36 & 92.99 & 73.86 \\ \hline
SAM2 Hiera  & Dice & 93.57 & \textbf{93.61} & \textbf{99.98} & 90.22 & \textbf{94.71} & 96.48 & \textbf{82.65} \\  \cline{2-9}
 Large      & IoU  & 89.31 & \textbf{90.39} & \textbf{99.96} & 83.89 & \textbf{90.68} & 93.98 & \textbf{75.93} \\
\hline
\end{tabular}
}
\end{table}

\begin{table}[t]
\centering
\caption{Comparison of Dice scores across various modalities for different selection strategies. R represents the number of reference frames.}
\label{tab:tab2}
\resizebox{\columnwidth}{!}{
\begin{tabular}{l|c|ccccccc}
\hline
Modality & R & Der & Endo & Fundus & OCT & US & XRay & MG \\
\hline
Random         & 1 & 59.84 & 59.77 & 99.95 & 31.99 & 99.69 & 42.88 & 46.34 \\
Uniform        & 5 & 50.52 & 52.41 & \textbf{99.97} & 30.65 & 99.83 & 37.01 & 43.24 \\
AFSE (wo scorer) & 5 & 62.54 & 50.83 & 99.67 & 23.24 & 99.76 & 55.79 & 39.85 \\ \hline
AFSE           & 5 & \textbf{62.55} & \textbf{60.22} & 99.75 & \textbf{33.90} & \textbf{99.98} & \textbf{60.07} & \textbf{50.82} \\
\hline
\end{tabular}
}
\end{table}

\begin{table}[t]
\centering
\caption{Comparison of Dice scores across different prompt strategies and medical imaging modalities.}
\label{tab:tab3}
\resizebox{\columnwidth}{!}{
\begin{tabular}{l|ccccccc}
\hline
Modality       & Der   & Endo  & Fundus & OCT   & US    & XRay  & MG    \\
\hline
Standard Pos   & 82.71 & 77.25 & 99.88  & 63.06 & 84.56 & 58.78 & 58.36 \\
Random Pos     & 74.36 & 81.37 & 99.82  & 57.48 & 79.62 & 52.21 & 53.11 \\
Single Neg     & 55.42 & 45.85 & 99.62  & 31.37 & 54.55 & 37.33 & 49.64 \\
Single Pos Neg & 82.41 & 50.94 & 99.78  & 61.86 & 83.29 & 63.48 & 58.49 \\
Four Pos       & 76.07 & 81.69 & 99.76  & 59.14 & 82.21 & 51.78 & 56.19 \\
Four Neg       & 54.43 & 00.78 & 99.61  & 34.75 & 50.11 & 37.54 & 44.86 \\
Single Pos Two Neg & 80.88 & 88.54 & 99.87 & 56.25 & 83.75 & 69.21 & 59.79 \\
Two Pos Four Neg   & 80.24 & 54.13 & 99.87 & 51.73 & 82.56 & 67.41 & 60.31 \\
\hline
BBox           & \textbf{94.79} & \textbf{85.27} & \textbf{99.92} & \textbf{90.31} & \textbf{94.44} & \textbf{96.55} & \textbf{81.56} \\
\hline
\end{tabular}
}
\end{table}

\subsection{Ablation Studies}

\textbf{Robust Prompts for SAM2 Propagation Segmentation.}  
In Table \ref{tab:tab3}, we compare the performance of various combinations of positive and negative point prompts with bounding box prompts across different medical imaging modalities. The results show that bounding box prompts, which enclose the target organ or lesion, consistently deliver the best segmentation performance by effectively leveraging SAM2's propagation mechanism. In contrast, the performance of point-based prompts is highly dependent on the correct combination of positive and negative points.

\textbf{Scoring Formula.}  
As demonstrated in Table \ref{tab:tab4}, AFSE effectively selects relevant frames for segmentation, significantly improving both efficiency and accuracy. By employing an unsupervised scoring mechanism that evaluates image features such as brightness, contrast, edge density, color histogram similarity, and shape similarity, AFSE automatically selects and ranks frames. This automated approach reduces the need for manual intervention, streamlining the segmentation process, while also maintaining robust generalization across diverse imaging modalities.

\begin{table}[H]
\centering
\caption{Comparison of Dice scores for different evaluation strategies across medical imaging modalities. AFSE represents a comprehensive evaluation mechanism that intelligently combines these metrics.}
\label{tab:tab4}
\resizebox{\columnwidth}{!}{%
\LARGE 
\begin{tabular}{l|ccccccc}
\hline
Modality         & Der   & Endo  & Fundus & OCT   & US    & XRay  & MG    \\
\hline
Brightness       & 62.31 & 50.41 & 99.67  & 22.71 & 99.78 & 55.79 & 39.38 \\
Color Histogram  & 62.71 & 51.67 & 99.67  & 23.24 & 99.76 & 55.79 & 39.85 \\
Contrast         & 62.08 & 51.13 & 99.67  & 23.24 & 99.78 & 55.79 & 39.38 \\
Edge Density     & \textbf{63.39} & 53.71 & 99.67  & 22.11 & 99.75 & 55.47 & 39.49 \\
Shape Similarity & 62.09 & 52.89 & 99.69  & 24.99 & 99.77 & 43.21 & 41.97 \\
\hline
AFSE             & 62.55 & \textbf{60.22} & \textbf{99.75} & \textbf{33.91} & \textbf{99.83} & \textbf{60.07} & \textbf{50.82} \\
\hline
\end{tabular}
}
\end{table}

\section{CONCLUSION}
In this work, we introduce the Strategy-driven Interactive Segmentation Model (SISeg), which enhances medical image segmentation across multiple modalities by integrating diverse prompt types. Using the Adaptive Frame Selection Engine (AFSE), SISeg dynamically selects optimal prompts without requiring prior medical knowledge, reducing memory usage and improving interpretability. Experiments on 10 datasets across 7 modalities show SISeg's ability to boost segmentation efficiency and lower annotation costs.

\section{ACKNOWLEDGMENTS}
This research is supported by the National Natural Science Foundation of China (No. 62201323, No. 62206242), and the Natural Science Foundation of Jiangsu Province (No. BK20220266).

\end{document}